%% file: main.tex
\documentclass[conference]{IEEEtran}
\usepackage{times}
\usepackage{amsmath}
\usepackage{amssymb}
\usepackage{tikz}
\usepackage[numbers]{natbib}
\usepackage{balance}
\usepackage{multicol}
\usepackage[bookmarks=true]{hyperref}
\usepackage{algorithm}
\usepackage{algorithmic}
\usepackage{cleveref}
\usepackage{booktabs}
\usepackage{threeparttable}
\usepackage[section]{placeins}
\usepackage{capt-of} 
\usepackage{utfsym}
\usepackage[font=small]{caption}
\usepackage{subcaption}

\begin{document}

\title{VRA: Grounding Discrete-Time Joint Acceleration in Voltage-Constrained Actuation}

\author{
\authorblockN{
Lingwei Zhang,
Jiaming Wang,
Tianlin Zhang, \\
Zhitao Song,
Xuanqi Zeng,
Weipeng Xia,
Zhongyu Li,
Yun-hui Liu$^{*}$
}
\authorblockA{
Department of Mechanical and Automation Engineering;
Hong Kong Embodied AI Lab, \\ The Chinese University of Hong Kong\\
Email: \{lwzhang, jmwang, tlzhang, ztsong, xqzeng, wpxia, zyli, yhliu\}@mae.cuhk.edu.hk\\
$^{*}$Corresponding author. Website and code: \href{https://zlwsss.github.io/VRA/}{https://zlwsss.github.io/VRA/}
}
}


%

\maketitle

\begin{abstract}
\input{content/abstract}
\end{abstract}

\IEEEpeerreviewmaketitle





\section{INTRODUCTION}
\input{content/Introduction}
\vspace{-0.2cm}
\section{RELATED WORK}
\input{content/related_work}
\section{PROBLEM SETUP}

\input{content/problem_setup} \label{sec:problem}
\section{PROPOSED METHOD}
\input{content/Proposed_Approach}

\section{EXPERIMENTAL RESULTS}

\input{content/Experiments}
\section{DISCUSSION AND LIMITATIONS}
\input{content/Discussion}

\section{CONCLUSION}
\input{content/Conclusion}
\section{ACKNOWLEDGMENT}
\input{content/acknowledgement}

\balance
\bibliographystyle{plainnat}
\bibliography{references}

\end{document}

%% file: content/abstract.tex
Discrete-time joint acceleration constraints are widely used to
enforce position and velocity limits. However, under
voltage-constrained electric actuators, kinematically admissible
accelerations may be physically unrealizable, exposing a missing
execution-level abstraction. We propose Voltage-Realizable
Acceleration (VRA), a joint-level acceleration interface that
grounds kinematic acceleration in voltage-constrained
actuator physics by restricting commanded accelerations to
voltage-realizable constraints. Hardware experiments on electric
actuators and a wheel-legged quadruped show that VRA removes
unrealizable accelerations, restores consistent near-constraint
execution, and reduces constraint-induced oscillations.

%% file: content/Introduction.tex
\begin{figure}[t]
    \centering
    \begin{subfigure}{\columnwidth}
        \centering
        \includegraphics[width=\linewidth]{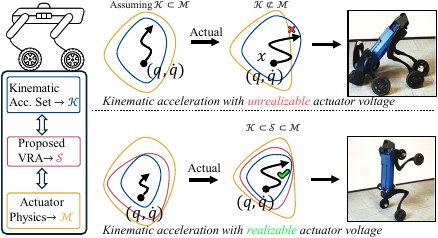}
        \caption{\textbf{Motivation}: kinematic realizability $\ne$ actuator voltage realizability}
        \label{fig: motivation}
        \vspace{0.5em}
    \end{subfigure}

    \begin{subfigure}{\columnwidth}
        \centering
    \includegraphics[width=0.8\linewidth]{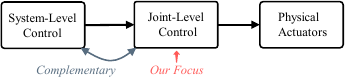}
        \caption{\textbf{Focus} of this work}
        \label{fig: focus}
    \end{subfigure}
    \caption{ \textbf{Motivation and scope of this work}. (a) System-level realizable motion commands may become unrealizable under actuator voltage constraints (top). Encoded with motor intent, VRA reshapes kinematic commands to remain executable. (b) We elevate joint-level control to explicitly address execution feasibility, which is implicitly assumed but not guaranteed by system-level controllers.}
    \label{fig: Overview}
\end{figure}

Discrete-time kinematic reasoning has become a standard tool for enforcing joint-level constraints in robot motion control. Many existing kinematic reasoning methods~\cite{PreteVK, Inverse_Kinematics_Redundant_Manipulators} generate admissible joint motions by propagating kinematic constraints in discrete time and are widely used as upstream modules for whole-body control~\cite{Zhang_2025_ViabilityIK, passive_Control, Robust_Walking}. Under this formulation, system-level feasibility is established at the kinematic level, while the execution of the resulting joint commands is delegated to lower-level controllers.
This separation implicitly assumes that joint-level commands deemed admissible by discrete-time kinematic reasoning can be reliably realized by the underlying actuators~\cite{underactuated}. However, as illustrated in Fig.~\ref{fig: Overview}(a), under voltage-constrained actuation, this assumption breaks down in two structural ways. First, kinematic admissibility is evaluated independently of actuator execution dynamics, so acceleration envelopes may include commands that are physically unrealizable at high joint speeds. Second, kinematic reasoning is performed in discrete time, while physical feasibility is determined by continuous-time actuator dynamics, leading to a mismatch between reasoning and execution.

\begin{figure*}[t]
  \centering
  \captionsetup{skip=2pt}
  \includegraphics[width=\textwidth]{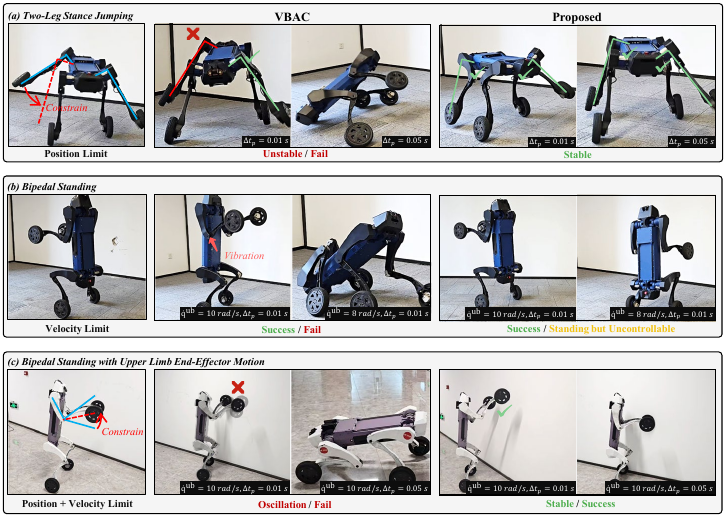}
  \caption{\textbf{Experiment snapshots under joint constraints}. Each row corresponds to one whole-body task, and columns consistently show the constraint setup, VBAC (commonly used in previous work\cite{passive_Control, PreteVK, Zhang_2025_ViabilityIK}), and the proposed method. Kinematically admissible but voltage-unrealizable accelerations can induce vibration or task failure near joint constraints, whereas the proposed method maintains more stable execution. All high-level whole-body controllers are unaware of the imposed joint constraints. Robots with different colors have the same actuators and hardware setup.}
  \label{fig: robot_result}
\end{figure*}

These mismatches arise from a missing abstraction between kinematic reasoning and joint-level execution. As shown in~Fig.~\ref{fig: Overview}(b), joint-level acceleration bounds form a fundamental interface for higher-level controllers. When this interface admits physically unrealizable accelerations, execution closure is broken on real hardware. Rather than addressing this issue at the whole-body controllers~\cite{MOR, MIT-human, SingleLeg}, this work focuses on the joint level and its role in mediating between kinematic reasoning and actuator limits.

In this paper, we propose Voltage-Realizable Acceleration (\textbf{VRA}), a method that explicitly connects actuator dynamics with kinematic reasoning. VRA operates at the joint acceleration level, providing kinematic reasoning with realizable acceleration bounds under voltage-constrained actuation and filtering out acceleration commands that are physically unrealizable due to model discrepancies. It treats kinematic constraints \textbf{solely} as \textbf{kinematic envelopes} that describe admissible motion ranges. The proposed method is a \textbf{precondition for any discrete-time kinematic reasoning method}. The kinematic admissibility at a discrete step does not imply physical realizability, and the assumed state transition underlying constraint reasoning may be invalidated at execution, even when all kinematic constraints are satisfied.

This paper makes the following contributions:
\begin{itemize}
    \item We identify a missing execution-level abstraction in discrete-time robot control: joint acceleration bounds that seem kinematically realizable can become physically unrealizable under voltage-constrained electric actuators.
    \item We propose VRA, a joint-level acceleration interface
    that bridges kinematic reasoning with actuator voltage constraints through acceleration bounds.
    \item We present the first systematic hardware study of discretization–actuation mismatch, showing that VRA eliminates unrealizable accelerations and restores consistency between kinematic reasoning and actuator execution with reduced boundary-induced oscillations.
\end{itemize}

Since this work focuses exclusively on joint-level acceleration, system-level constraint satisfaction methods such as Control Barrier Functions (CBF)~\cite{CBF_ORIGIN, CBF_WALK, CBF_WALK2,CBF_Robust_WALK} and hierarchical quadratic programs~\cite{Inverse_Kinematics_Redundant_Manipulators,Efficient_Paradigm,feasible_region} fall outside the scope of this paper.

%% file: content/related_work.tex
\textbf{Viability-based joint-level control} explicitly reasons about the existence of future feasible trajectories under state and input constraints, providing forward invariance and recursive feasibility guarantees. Representative work~\cite{PreteVK} derives state-dependent joint acceleration bounds to ensure satisfaction of joint position and velocity limits in discrete time, and has been extended to a multi-joint QP framework~\cite{Zhang_2025_ViabilityIK}, inverse kinematics on manipulators~\cite{Inverse_Kinematics_Redundant_Manipulators}, and passive torque control~\cite{passive_Control}. While these methods provide strong theoretical guarantees, they rely on discrete-time kinematic propagation and often state-independent acceleration limits. As a result, feasibility is defined over future trajectories rather than instantaneous physical execution, which can lead to a mismatch when deployed on voltage-constrained hardware.

\textbf{Joint-level actuator-aware control.} Existing joint-level actuator-aware methods address actuator limits primarily through post-hoc handling of infeasible commands. Representative approaches clip torque commands onto fixed admissible regions after high-level commands are generated~\cite{MIT-human,MOR,SingleLeg}. Input saturation control methods focus on maintaining stability after actuator saturation has already occurred, treating saturation as an execution-side effect~\cite{Input_sat}. In both cases, feasibility is enforced at the command level and remains decoupled from kinematic constraints. In contrast, our approach is also post-hoc but differs fundamentally in where and how post-hoc handling is applied. 
VRA operates at the joint acceleration level before torque saturation occurs and explicitly filters physically unrealizable acceleration commands under voltage-constrained actuation. Rather than modifying commands internally within a controller, VRA exposes voltage-feasible acceleration bounds as an explicit execution interface that higher-level controllers can reason about. This makes the post-hoc effect visible upstream and couples kinematic envelopes with actuator realizability at execution time. 

\begin{table}[t]
  \centering
  \scriptsize
  \setlength{\tabcolsep}{3pt}
  \renewcommand{\arraystretch}{0.8}
  \begin{threeparttable}
  \caption{Relationship of Joint-Level Methods}
  \label{tab:relatedwork}
  \begin{tabular}{lcccc}
    \toprule
    \textbf{Method}
    & Post-hoc
    & Post-hoc Target
    & Voltage Realizable
    & Viable \\
    \midrule
    Zhang et al.~\cite{Zhang_2025_ViabilityIK} & $\checkmark$ & Feasible Set & $\usym{2717}$ & $\checkmark$ \\
    Wen et al.~\cite{Input_sat} & $\checkmark$ & Torque Command & $\usym{2717}$ & $\usym{2717}$ \\
    Kang et al.~\cite{SingleLeg} & $\checkmark$ & Torque Command & $\usym{2717}$ & $\usym{2717}$ \\
    Morimoto et al.~\cite{motor_well_known_1} & $\usym{2717}$ & - & $\usym{2717}$ & $\usym{2717}$ \\
    \midrule
    VRA (Ours) & $\checkmark$ & Acceleration Set & $\checkmark$ & $\usym{2717}$ \\
    \bottomrule
  \end{tabular}
  \end{threeparttable}
\end{table}

\textbf{Voltage-constrained motor physics.} In motor control literature, voltage-constrained motor physics and the resulting torque–speed envelopes are well understood~\cite{motor_well_known_1,motor_well_known_2}, and control methods primarily focus on anti-saturation~\cite{MOTOR_ANTI_SATURATION, MOTOR_ANTI_SATURATION_1} and on maintaining stability and performance during saturation~\cite{Motor_SAT_POLICY}. These methods operate at the drive level, where saturation is treated as an operating condition rather than a constraint anticipated by high-level motion generation, leading robot-level control to abstract voltage limits into static torque bounds and leaving a gap with acceleration-level motion reasoning. 

As shown in Fig.~\ref{fig: method_frame}, our proposed method does not constitute a safety guarantee and does not imply invariance or recursive feasibility. Instead, it defines a minimal realizability contract required for kinematic acceleration constraints to be meaningful at execution time. Nominally, VRA may appear as a post-hoc strategy, as physically unrealizable joint acceleration commands are clipped before execution. However, this post-hoc behavior is not intrinsic to the method itself. VRA explicitly provides kinematic reasoning modules with realizable joint acceleration bounds derived from actuator dynamics, enabling upstream controllers to operate within physically feasible acceleration envelopes. Post-hoc clipping only arises in practice due to model discrepancies, rather than from the design principle of VRA. We summarize the relationship of this work with existing methods in Table~\ref{tab:relatedwork}.

%% file: content/problem_setup.tex
We consider a robotic system operating under discrete-time joint-level acceleration commands. Throughout this paper, acceleration bounds are treated as an interface variable between kinematic reasoning and actuator execution, rather than as a safety guarantee. The single joint kinematics set $\mathcal{F}$ is defined as:
\begin{equation}
     \mathcal{F}=\{(q,\dot{q}) \in \mathbb{R}^{2}:{q}^\mathrm{lb}< q<{q}^\mathrm{ub},|\dot{q}|\leq  \dot{q}^\mathrm{ub}\},
\end{equation}
where $q$ is the joint position, $\dot{q}$ is the joint velocity, $(\cdot^{\mathrm{lb}})$ denotes the lower bound, $(\cdot^{\mathrm{ub}})$ denotes the upper bound. The set is approximated in discrete time by restricting the commanded joint acceleration to
\begin{equation}
    \ddot{q}_k \in[\ddot{q}_k^\mathrm{lb},\ddot{q}_k^\mathrm{ub}], \label{eq: acceleration_limits}
\end{equation}
where $k$ denotes the discrete-time step. In this work,~\eqref{eq: acceleration_limits} are used solely as discrete-time kinematic acceleration limits, serving as the input language of higher-level controllers, representing a geometric envelope on admissible accelerations rather than guaranteeing invariance or recursive feasibility.

The kinematic admissibility defined above does not specify whether the underlying actuators can physically realize a commanded acceleration at execution time. To characterize execution-level feasibility, we introduce a voltage-constrained motor model and explicitly distinguish
between kinematic admissibility and physical realizability.

Joint accelerations are generated by joint torques $\boldsymbol{\tau}_k$, which are produced by electric motors through a transmission. We assume each joint is driven by a voltage-constrained electric motor. In the motor $dq$ reference frame, the applied voltage $\boldsymbol{u}_k=[u_{d,k}\;u_{q,k}]^\top$ satisfies
\begin{equation}
    \|\boldsymbol{u}_k(i_{q,k},\dot{q}_k)\|\leq V_{\mathrm{bus}},
\end{equation}
where $i_{q,k}$ is the $q$-axis current, $V_{\mathrm{bus}}$ is the DC-link voltage.

Under the standard assumption, the $d$-axis current $i_{d,k}=0$ (flux weakening is out of scope~\cite{minicheetah, MIT-human}), the motor voltage is given by
\begin{equation}
\label{eq:motor_voltage}
    \begin{bmatrix}
        u_{d,k} \\
        u_{q,k}
    \end{bmatrix}
    =
    \begin{bmatrix}
        -p\dot{q}_k L_q i_{q,k} \\
        R_s i_{q,k} + p\dot{q}_k\varphi
    \end{bmatrix},
\end{equation}
where $L_q, R_s,$ and $\varphi$ are motor parameters,  $p$ is a constant convert joint velocity to electric velocity. As a result, the feasible motor torque—and hence the feasible joint acceleration—is state dependent and decreases with increasing joint velocity.

\begin{figure}[!t]
\centering
\includegraphics[width=3.5in]{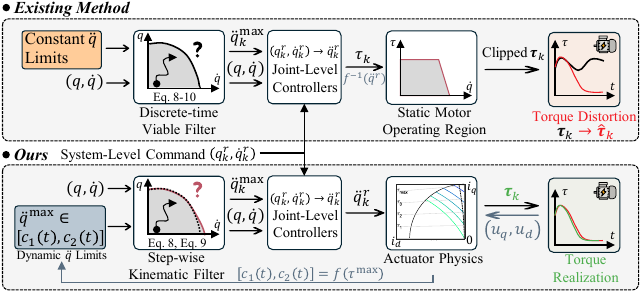}
\caption{\textbf{Overview of the proposed method}. Existing methods~\cite{Zhang_2025_ViabilityIK, PreteVK, passive_Control} reason over kinematics with constant braking limits and rely on post-hoc clipping~\cite{MOR,SingleLeg} to handle actuator saturation (top). In contrast, our method connects kinematic reasoning and actuator dynamics through VRA (bottom). ${(\cdot^r)}:$ denotes references, $\hat{(\cdot)}$ denotes actual effort.}
\label{fig: method_frame}
\end{figure}

The problem addressed in this paper is to characterize, at each discrete time step, a joint-level method interface that maps kinematically admissible joint accelerations to those that are physically realizable under voltage-constrained actuation. Specifically, an acceleration is said to be realizable under kinematic constraints and actuator dynamics if there exists a motor current $i_{q,k}$ such that
\begin{equation}
\begin{aligned}
        &\tau_k=k_ti_{q,k}, \\
        &\tau_k=M\ddot{q}_k+h(\dot{q}_k), \\
        &\Vert\boldsymbol{u}_k(i_{q,k},\dot{q}_k)\Vert \leq V_{\mathrm{max}}, \\ &\ddot{q}_k\in[\ddot{q}_k^{\mathrm{lb}},\ddot{q}_k^{\mathrm{ub}}], \\
        & (q_k,\dot{q}_k) \in \mathcal{F}. \\
\end{aligned}
\end{equation}
where $k_t$ is a constant, $M$ is a scalar inertia for single joint, $h(\dot{q}_k)$ collects friction and viscous resistance. 
These constraints induce a state-dependent feasible torque interval
\begin{equation}
    \tau_k \in [\tau_k^{\mathrm{lb}}(\dot q_k),\;\tau_k^{\mathrm{ub}}(\dot q_k)],
\end{equation} within which any torque corresponding to an admissible acceleration
is guaranteed to be realizable.

\noindent\textbf{Remark 1 (Structural gap of Discrete-Time Kinematic Reasoning under Voltage Limits).}
This formulation highlights that acceleration feasibility is governed by actuator physics and is inherently state-dependent under voltage limits. In contrast, kinematic reasoning derives admissible accelerations from discrete-time propagation, yielding bounds that scale with $1/\Delta t$ and are intrinsically unbounded. Excessively small $\Delta t$ leads to physically unrealizable acceleration demands, while large $\Delta t$ induces over-conservative execution.

%% file: content/Proposed_Approach.tex
\subsection{Acceleration Realizability from Voltage Constraints} \label{sec:VATS}
We connect kinematic reasoning with actuator dynamics by deriving joint acceleration bounds from actuator voltage constraints. The voltage constraint directly restricts the set of realizable joint accelerations, and the resulting acceleration bounds serve as the coupling between actuator voltage realizability and kinematic reasoning.

The proposed construction is general to PMSM-based quasi-direct-drive electric actuators. It does not assume a particular low-level current controller or modulation strategy, but only uses the voltage finally applied to the motor. This voltage can be measured from the drive or reconstructed from the motor state. Together with the motor current, joint speed, and applied voltage, it is sufficient to evaluate the voltage-realizable acceleration set. In this work, we consider the standard case $i_d=0$, while the field-weakening operation is left outside the scope.

We rely on two mild assumptions to enable real-time evaluation of acceleration bounds from voltage constraints: 
\begin{itemize}
    \item Electrical-mechanical time-scale separation: The electrical speed is frozen over one control step, exploiting the separation between fast electrical and slower mechanical dynamics. This approximation reduces the voltage feasibility condition to a one-dimensional quadratic constraint that can be evaluated in closed form at each step.
    \item Bounded model discrepancies: Unmodeled effects between the nominal motor model $\boldsymbol{u}_{k}^\mathrm{n}(\omega_{e,k}, i_{q,k})$ and the actual actuator behavior (such as DC-link ripple, dead time, and parameter mismatch~\cite{DeltaVMotor}) are lumped into a residual voltage term $u^{\mathrm{res}}_k$. This term eliminates model drift by successively compensating for unmodeled effects, making voltage feasibility valid across all operating conditions.
\end{itemize}

Under these assumptions, voltage realizability is approximated through two complementary constraints: a transient voltage-realizable set $\mathcal{S}_{\mathrm{tr}}$ and a quasi-steady look-ahead set $\mathcal{S}_{\mathrm{st}}$. These two sets both form a quadratic inequality:
\begin{equation}\label{Eq:quadratic}
    \mathcal{S}_{i,k} \triangleq \{ i_{q,k} \mid  A_i (i_{q,k})^2 + B_i i_{q,k} + C_i \le 0 \},i\in\{\mathrm{tr},\mathrm{st}\},
\end{equation}
where $A_i,B_i,C_i$ are state-dependent parameters at control step $k$. 
Both sets are necessary: the transient set $\mathcal{S}_{\mathrm{tr}}$ guarantees instantaneous voltage realizability under discrete-time current dynamics, while the quasi-steady set $\mathcal{S}_{\mathrm{st}}$ provides a look-ahead constraint that anticipates near-future back-EMF effects. With $i_{q,k}$ linearly proportional to $\tau_k$, the intersection set $\mathcal{I}$ of the two voltage-realizable sets yields a state-dependent admissible torque interval, which serves as the auxiliary set of the acceleration set. Detailed derivations and implementation specifics are provided in the supplementary material. 

\noindent \textbf{Lemma~1~(Acceleration Realizability under Voltage Constraints)}:
At each discrete-time step $k$, any joint acceleration command $\ddot{q}_k$ produced by VRA admits an acceleration realizability under the actuator voltage constraint $\Vert u_k\Vert \leq V_{\mathrm{max}}$, assuming the actuator model and sampling conditions stated in the two assumptions.

\noindent \textbf{Proof}: VRA explicitly constructs, at each step $k$, the set of admissible $q$-axis currents as
\[
\mathcal{I}_k=\mathcal{S}_{\mathrm{tr},k} \cap \mathcal{S}_{\mathrm{st},k},
\]
where both sets are derived directly from the discrete-time motor voltage constraint $\Vert u_k\Vert\leq V_{\mathrm{max}}$. By construction, any current $i_{q,k}\in \mathcal{I}_k$ satisfies the voltage limit under the assumed motor dynamics. Since joint torque is linearly related to $i_{q,k}$, any acceleration whose realizability requires $i_{q,k}\in \mathcal{I}_k$ admits a realizable torque. VRA enforces this by selecting the commanded acceleration within the set induced by $\mathcal{I}_k$. Therefore, the acceleration command produced by VRA is realizable under the motor voltage constraint at step $k$.

Lemma 1 establishes that VRA guarantees instantaneous acceleration realizability under voltage constraints by construction.

\subsection{Acceleration Realizability from Kinematic Constraints}
We directly follow the kinematic envelopes in~\cite{PreteVK}. The acceleration constraints from position limits, 
\begin{equation}
    {q}^{\mathrm{lb}}\leq q+\dot{q}\ t+0.5\ddot{q}\ t^2 \leq q^{\mathrm{ub}}, \quad \forall t\in[0,\Delta t_p],
\end{equation}
where $\Delta t_p$ is the discrete-time step in kinematic reasoning, and the acceleration constraints from the position-velocity relationship, 
\begin{equation}
    \dot{{q}}_{k+1}\leq \sqrt{2\ddot{q}^\mathrm{lb}_k(q^{\mathrm{lb}}-{q}_k-\Delta t_p \dot{{q}_k}-0.5\Delta t_p^2 \ddot{q}_k)},
\end{equation}
only become active when states approach the position boundary. Unlike position-related constraints, the discrete-time velocity constraints, 
\begin{equation}
    \frac{1}{\Delta t_p}(-\dot{q}^\mathrm{ub}-\dot{q})\leq \ddot{q} \leq \frac{1}{\Delta t_p}(\dot{q}^\mathrm{ub}-\dot{q}),
\end{equation}
induces acceleration bounds that scale with $1/\Delta t_p$ and can affect performance far from the boundary. The acceleration bounds induced by the discrete-time kinematic constraints above are insufficient to characterize realizable acceleration during motor actuation. The realizable deceleration exhibits a non-monotonic dependence on joint speed that cannot be captured by constant-acceleration bounds and discrete-time kinematic propagation.

This behavior arises from actuator dynamics in~\eqref{eq:motor_voltage}. Near velocity bounds, discrete-time kinematic reasoning may allow large deceleration commands, but implementing them requires high negative $q$-axis current, which induces strong speed-dependent $dq$-axis voltage coupling that rapidly exhausts the available voltage budget. As a result, achievable deceleration weakens at high speed. As the actuator slows down, the coupling diminishes, the voltage margin is released, and stronger deceleration becomes realizable. This inherent weak–then–strengthen deceleration behavior is a direct consequence of coupled $dq$-axis voltage constraints and is fundamentally incompatible with constant acceleration limits.

At the velocity level, the actuator dynamics manifest as explicit voltage constraints. In an electric actuator without field-weakening control, to avoid degrading performance, the maximum joint velocity is typically set above the motor corner speed. In this high-speed region, the back-EMF dominates the voltage balance, with the q-axis current being relatively small. Consequently, the current-dependent terms in \eqref{eq:motor_voltage} can be neglected compared to the EMF term, and the achievable maximum speed is directly proportional to the available voltage budget:
\begin{equation}
    u_q\approx \omega_e\varphi \quad \Rightarrow \dot{q}^{\mathrm{max}}\propto V_{\mathrm{limit}}.
\end{equation}
Consequently, the maximum admissible velocity is structurally determined by the voltage constraint itself and can be adjusted by scaling the voltage budget $V_{\mathrm{limit}}$.  

This observation is important because the velocity bound is
not introduced as an independent discrete-time constraint. Instead,
it is absorbed into the same voltage constraints that
define acceleration realizability. As the actuator approaches the
voltage boundary, the feasible current interval naturally shrinks,
and the realizable acceleration set continuously reflects the
remaining voltage margin.

With velocity limits encoded through motor voltage feasibility, the resulting velocity boundary is less conservative, as it arises from continuous-time motor physics rather than discrete-time propagation. 

\subsection{Acceleration Realizability from Actuation and Kinematics}
In the discrete-time setting, at step $k$, the acceleration bounds induced by voltage constraints are computed from the measured state history as follows:
\begin{subequations}
\begin{align}
    u_{d,k-1}^\mathrm{n} &:= -p\dot{q}_{k-2}i_{q,k-1}L_q \\
    u_{q,k-1}^\mathrm{n}&:=R_si_{q,k-1}+p\dot{q}_{k-2}\varphi \\
    u_{d,k}^\mathrm{res}&:=u_{d,k-1}^\mathrm{n}-u_{d,k-1}\\
    u_{q,k}^\mathrm{res}&:=u_{q,k-1}^\mathrm{n}-u_{q,k-1}\\
    \hat{u}_{d,k}&:=-p\dot{q}_{k-1}i_{q,k}L_q+u_{d,k}^\mathrm{res} \\
    \hat{u}_{q,k}^{\mathrm{tr}}&:=i_{q,k}R_s+\frac{i_{q,k}-i_{q,k-1}}{\Delta t}p\dot{q}_{k-1}\varphi+u_{q,k}^{\mathrm{res}} \\
    \hat{u}_{q,k}^{\mathrm{st}}&:=i_{q,k}R_s+(\dot{q}_{k-1}+\frac{k_ti_{q,k}}{M}s\Delta t)p\varphi+u_{q,k}^{\mathrm{res}} \\
    \mathcal{S}_{\mathrm{tr},k}&:=\{i_{q,k}|\sqrt{\hat{u}_{d,k}^2+(\hat{u}_{q,k}^\mathrm{tr})^2}\leq V_\mathrm{limit}\}\\
     \mathcal{S}_{\mathrm{st},k}&:=\{i_{q,k}|\sqrt{\hat{u}_{d,k}^2+(\hat{u}_{q,k}^\mathrm{st})^2}\leq V_\mathrm{limit}\}\\
     \mathcal{I}_k&:=\mathcal{S}_{\mathrm{tr},k}\cap \mathcal{S}_{\mathrm{st},k}\\
     \mathcal{A}_{v,k} &:=M^{-1}k_t\mathcal{I}_k-M^{-1}h(\dot{q}_k).
\end{align}
\end{subequations}

All computations are carried out with a discrete-time step $\Delta t$. The parameter $s$ in~(12g) specifies the look-ahead step and is typically chosen to be greater than the communication delay of the robotic system. 

As the actuator approaches its voltage limit, the admissible $q$-axis current $i_{q,k}$ tends to zero. In this regime, the voltage constraints in~(12h) and~(12i) degenerate to
\[
p\,\dot q_{k-1}\,\varphi \leq V_{\mathrm{limit}},
\]
which directly imposes an upper bound on the achievable joint velocity. Consequently, the maximum admissible speed can be regulated by tuning the available voltage budget
$V_{\mathrm{limit}}$. With velocity limits inherently encoded by motor voltage feasibility, the discrete-time velocity constraints in the kinematic envelope become unnecessary and can be removed. Then the acceleration bounds from kinematic constraints can be computed as: 
\begin{subequations}
\begin{align}
\ddot{q}_k^\mathrm{max}&:=\mathrm{inf}(\mathcal{A}_{v,k}),\mathrm{if}\  \dot{q}_k\geq0 \\
\ddot{q}_k^\mathrm{max}&:=\mathrm{sup}(\mathcal{A}_{v,k}),\mathrm{if}\  \dot{q}_k<0 \\
\mathcal{A}_{1,k} &:=\{\ddot{q}_k|\ddot{q}_k \in \mathrm{Algorithm1}(\dot{q}_{k-1},q_{k-1},\Delta t_p)\text{\cite{PreteVK}}\}\\
\mathcal{A}_{2,k}&:=\{\ddot{q}_k|\dot{q}_k+\Delta t_p\ddot{q}_k\leq\sqrt{2\ddot{q}_k^\mathrm{max}(q^\mathrm{ub}-q_k)}\}\\
\mathcal{A}_{3,k}&:=\{\ddot{q}_k|\dot{q}_k+\Delta t_p\ddot{q}_k\geq -\sqrt{2\ddot{q}_k^\mathrm{max}(-q^\mathrm{lb}+q_k)}\}\\
\mathcal{A}_k&:=\mathcal{A}_1\cap\mathcal{A}_2\cap\mathcal{A}_3
\end{align}
\end{subequations}

Since $\mathcal{A}_{v,k}\in\mathbb{R}$ is a closed interval induced by voltage feasibility, $\mathrm{sup}(\cdot)$ and $\mathrm{inf}(\cdot)$ are well-defined scalars, $\ddot{q}^{\mathrm{max}}_k$ is updated deceleration bounds. Kinematic reasoning is performed in discrete time with a step size $\Delta t_p$,
which is intentionally decoupled from the execution time step $\Delta t$.
This separation enables kinematic constraints to be reasoned over a look-ahead predictor~\cite{look_ahead} independent of actuator execution dynamics, while preserving execution
feasibility through the actuator-aware acceleration bounds.
Given the acceleration sets derived from actuator dynamics and from the kinematic
envelope, the commanded acceleration is required to lie in the intersection:
\begin{equation}
    \mathcal{A}_{\mathrm{cmd},k}:=\mathcal{A}_{v,k}\cap\mathcal{A}_k
\end{equation}

Please note that acceleration commands satisfying the set
$\mathcal{A}_{{\mathrm{cmd},k}}$ do not necessarily guarantee
actuator-level acceleration feasibility due to model
discrepancies in~(12k). Realizability is instead ensured by
the set $\mathcal{I}_k$. Therefore, the commanded acceleration must be projected back onto $\mathcal{I}_k$. The computational order of the proposed method is summarized as
\[
\mathrm{States}\rightarrow \mathcal{I}_k\rightarrow\mathcal{A}_{v,k}\rightarrow\mathcal{A}_k\rightarrow\mathcal{A}_{\mathrm{cmd},k}\rightarrow \mathcal{I}_k \rightarrow\mathrm{Actuators}.
\]
This order is fixed and essential. The actuator feasibility set $\mathcal{I}_k$ cannot encode position limits, while the
kinematic acceleration set $\mathcal{A}_k$ cannot capture motor voltage feasibility.
By computing $\mathcal{I}_k$ first, actuator voltage limits are translated into state-dependent acceleration bounds prior to kinematic reasoning, preventing kinematically realizable yet physically unrealizable commands. The final projection back to $\mathcal{I}_k$ ensures execution
realizability under model discrepancies and reflects a post-hoc adjustment necessitated by actuator physics rather than by the design principle itself.

%% file: content/Experiments.tex
Although VRA operates purely at the acceleration level, the experiments are designed to expose the mismatch between discrete-time kinematic acceleration envelopes and physically realizability imposed by actuator dynamics. Accordingly, we report induced torque and voltage trajectories not as control objectives, but as execution-level voltage realizability diagnostics that reveal whether kinematic acceleration commands admit physical actuator implementations.
\subsection{Experimental Setup}
We evaluate the proposed method on actuated joints, which isolates actuator-level effects while preserving the discrete-time control structure typically used in whole-body control. Unless otherwise stated, our method is executed with a fixed control period of $\Delta t=0.001$~s. The discrete-time used in kinematic reasoning is varied as $\Delta t_p\in\{0.001,0.005,0.01\}$~s. For comparison, the viability-based acceleration control~(VBAC)~\cite{PreteVK} is treated as a classical discrete-time joint-level kinematic control method, determining admissible joint accelerations purely from discrete-time propagation of position and velocity constraints. Independently, the motor operating region (MOR) is used as a standard actuator-level voltage-realizability clipping derived from the dynamometer~\cite{MIT-human,MOR}. The MOR is not combined with VBAC, but serves solely as a baseline actuator voltage-realizable set for evaluating how well different acceleration bounds reflect true actuator voltage realizability. Across all comparisons, we maintain the actuator condition identically, including reference commands, power-supply settings, and low-level torque control loops. Motor and drive setups are summarized in the supplementary materials.

\subsection{Validating Voltage-Realizable Acceleration from Actuator Constraints}
The objective of this experiment is to evaluate whether the proposed realizable q-axis current set accurately characterizes the actuator's voltage realizability. In particular, we examine whether the actuators can realize acceleration commands derived from the proposed current set without inducing voltage saturation across different motors and operating conditions.

Experiments are conducted on two motors with distinct inertial and electrical characteristics (Motor~8115 and Motor~6210). Each motor is commanded with aggressive step torque inputs (30~Nm/20~Nm) under nominal (24~V/25~$^\circ$C) and stressed (36~V/45~$^\circ$C) conditions. For each configuration, all methods are evaluated under identical actuator, power-supply, and low-level control settings. Following \cite{MOR,MIT-human}, MOR is constructed as a speed-dependent linear torque boundary identified at 48~V. For experiments under different bus voltages, the MOR is uniformly scaled according to the available voltage budget, ensuring a fair comparison across methods. Implementation details are provided in the supplementary material.

\begin{figure}[t]
\centering
\includegraphics[width=3.2in]{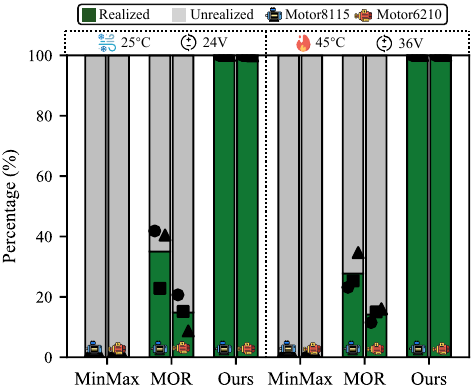}
\caption{\textbf{Proportion of realizable and unrealizable actuator voltage across methods and operating conditions.} Green and gray bars indicate realizable and unrealizable executions, respectively. Results are shown for different motors and voltage–temperature conditions, bars report the mean over trials, and markers indicate independent experimental repeats.} 
\label{fig:exection}
\end{figure}

\begin{figure*}[t]
    \centering
    \begin{minipage}{0.49\textwidth}
        \centering
        \includegraphics[width=\linewidth]{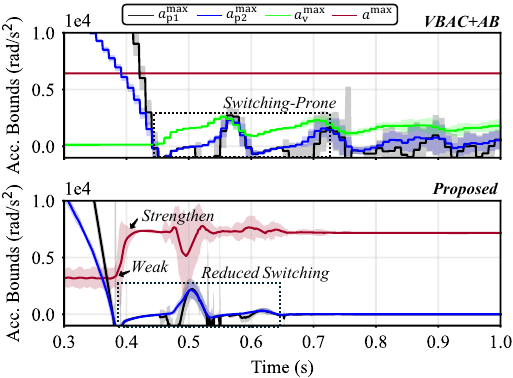}
        \small (a) Acceleration bounds comparison.
    \end{minipage}
    \hfill
    \begin{minipage}{0.49\textwidth}
        \centering
        \includegraphics[width=\linewidth]{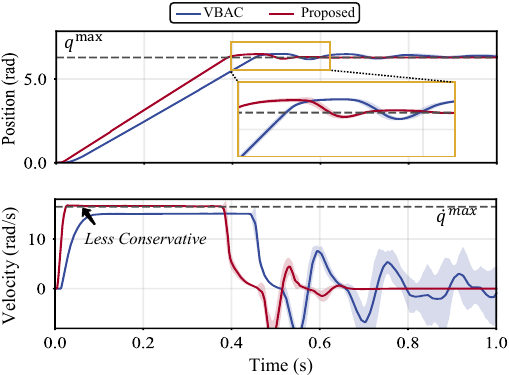}
        \small (b) Joint kinematics comparison.
    \end{minipage}
    \caption{\textbf{Best-case kinematic response with $\Delta t_p=0.01$~s}. Left: acceleration bounds, $a_{\mathrm{p_1}}^{\mathrm{max}}$ denotes the bounds from position constraints, $a_\mathrm{p_2}^{\mathrm{max}}$ denotes the bound from position-velocity constraints, $a_{\mathrm{v}}^\mathrm{max}$ denotes the bounds from velocity, $a^\mathrm{max}$ denotes the limits of acceleration. Right: motor position and velocity over the full motion toward the maximum position. Lines show the mean over 10 trials, with shaded regions denoting the $\pm$ std.} 
    \label{fig: Experiment_2}
    \vspace{-6pt}
\end{figure*}

\begin{figure}[t]
\centering
\includegraphics[width=3.2in]{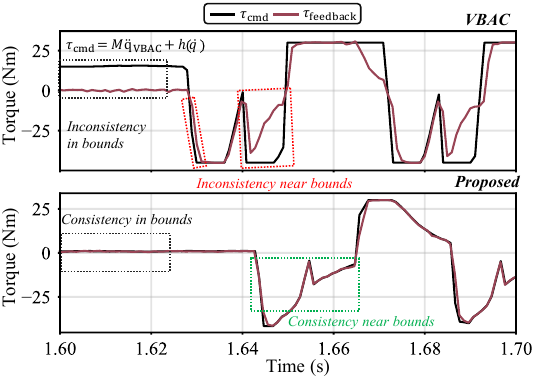}
\caption{\textbf{Kinematic feasibility does not imply execution feasibility}. Significant discrepancies in VBAC arise near the kinematic bounds, while the proposed method maintains torque consistency both inside and near the bounds.} 
\label{fig: exp2_explain}
\end{figure}

Fig.~\ref{fig:exection} reports the proportion of trials exhibiting voltage
saturation across methods and operating conditions. Across
both motors and all tested regimes, commands derived from
kinematic baselines frequently induce voltage saturation,
indicating a mismatch between kinematic admissibility and
actuator feasibility. In contrast, the proposed method
substantially reduces saturation events by explicitly
restricting commands to the admissible q-axis current set.

The underlying cause of voltage infeasibility is analyzed in the supplied materials. These results show that $\mathcal{I}_k$ captures voltage-realizable actuation more accurately than MOR. Unlike post-hoc saturation handling, VRA exposes voltage-realizable acceleration bounds upstream before commanding, with only residual clipping arising from model mismatch.

\subsection{Combining Kinematic Envelopes with Voltage-Realizable Acceleration}\label{label: test}
This experiment evaluates whether the proposed method can respect discrete-time kinematic constraints while operating strictly within voltage-realizable acceleration bounds, thereby validating the compatibility between actuator-aware acceleration
limits and kinematic reasoning.

\begin{figure*}[!t]
    \centering
    \begin{minipage}{0.49\textwidth}
        \centering
        \includegraphics[width=\linewidth]{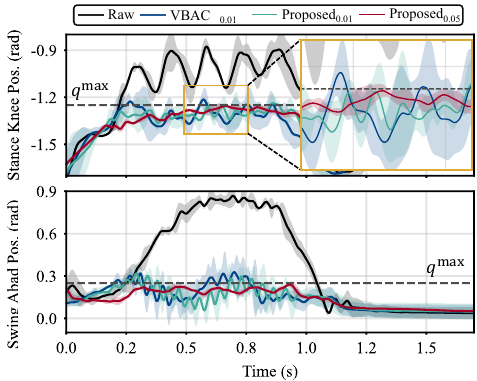}
        \small (a) Joint position comparison in jumping motion.
    \end{minipage}
    \hfill
    \begin{minipage}{0.49\textwidth}
        \centering
        \includegraphics[width=\linewidth]{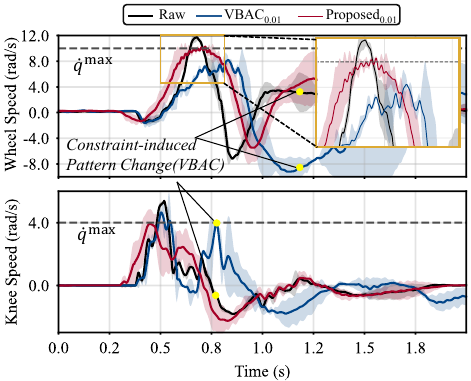}
        \small (b) Joint velocity comparison in bipedal standing.
    \end{minipage}
    \caption{\textbf{Joint-level execution behavior under whole-body controllers}. (a) Joint position trajectories during a two-leg jumping motion with position limits imposed on the swing and stance legs. (b) Joint velocity trajectories during bipedal standing with maximum velocity limits imposed on all actuated joints. The boundaries generated by VBAC induce qualitative changes in execution patterns, whereas the proposed method exhibits logically consistent behavior across both tasks. Solid lines indicate the mean over trials, and shaded regions denote one standard deviation. Raw denotes the original controller, and the corner label denotes the discrete-time in kinematic reasoning.}
    \label{fig:Experiment_3}
    \vspace{-6pt}
\end{figure*}

Based on the actuator-level consistency established in Experiment~1, this experiment is conducted on the single motor~8115 to isolate kinematic effects. With a fixed control period of $\Delta t=0.001$~s and a step acceleration command with realized torque (30~Nm), the joint is driven toward its position and velocity limits under a high-speed decelerating scenario. The proposed method is compared against VBAC baselines under identical conditions. After an initial parameter sweep, the best-performing discrete-time configuration is selected for each method, and the experiment is repeated ten times using this fixed setting.

Performance is evaluated over two time windows around the kinematic boundary, quantifying velocity tracking conservation, braking smoothness, and oscillatory behavior. A trial is marked realizable if commanded accelerations remain realizable during braking.

\begin{table}[t]
  \centering
  \scriptsize
  \setlength{\tabcolsep}{4pt}
  \renewcommand{\arraystretch}{0.95}
  \begin{threeparttable}
  \caption{Near-boundary Performance Metrics Comparison.}
  \label{tab:jitter}
  \begin{tabular}{lcccccc}
    \toprule
    \textbf{Method}
    & $\Delta t_p$
    & \textbf{Success}
    & $\Delta \dot q_{\max\downarrow}$
    & $|\Delta \dot q_{\mathrm{RMS}\downarrow}|$
    & $|\Delta \tau_{\mathrm{RMS}\downarrow}|$
    & $f_{\mathrm{zc}\downarrow}$ \\
    \midrule
    VBAC + AB & 0.001 & $\usym{2717}$ & -- & -- & -- & -- \\
    VBAC + CB  & 0.001 & $\checkmark$ & 1.1135 & 1.2503 & 21.6034 & \textbf{10.7} \\
    VRA (Ours) & 0.001 & $\checkmark$ & \textbf{0.0193} & \textbf{0.8890} & \textbf{8.5347} & 12.7 \\
    \midrule
    VBAC + AB & 0.005 & $\usym{2717}$ & -- & -- & -- & -- \\
    VBAC + CB  & 0.005 & $\checkmark$ & 0.1432 & 1.1585 & 18.1467 & 17.3 \\
    VRA (Ours) & 0.005 & $\checkmark$ & \textbf{0.0205} & \textbf{0.4825} & \textbf{4.7935} & \textbf{7.3} \\
    \midrule
    VBAC + AB & 0.01 & $\checkmark$ & 0.2793 & 0.4037 & 4.2471 & 19.3 \\
    VBAC + CB  & 0.01 & $\checkmark$ & 1.5162 & 0.1850 & 3.9275 & 18.3 \\
    VRA (Ours) & 0.01 & $\checkmark$ & \textbf{0.0173} & \textbf{0.1676} & \textbf{1.4670} & \textbf{4.7} \\
    \bottomrule
  \end{tabular}
  \begin{tablenotes}[flushleft]
    \footnotesize
    \item Values are averaged over time windows and across three trials. Units are: 
    $\Delta t_p$~(s), $f_{\mathrm{zc}}$~(Hz), $\dot{q}$~(rad/s), and $\tau$~(Nm). Success indicates whether commanded accelerations remain realizable during deceleration. AB and CB denote aggressive and conservative constant acceleration bounds, respectively.
  \end{tablenotes}
  \end{threeparttable}
\end{table}

Table~\ref{tab:jitter} summarizes joint-level metrics near kinematic boundaries. Rather than indicating tracking performance, these metrics quantify the stability of the execution interface, including induced oscillations and realizability during braking. Fig.~\ref{fig: exp2_explain} reveals the underlying cause: accelerations deemed admissible by discrete-time kinematic reasoning are mapped, via the nominal dynamics model, to torque realizations that violate voltage feasibility near the boundary, leading to torque distortion. In contrast, the proposed method maintains torque-realizable behavior with reduced oscillations, and reduces the discretization-induced speed conservation while respecting the kinematic constraints (See Fig.~\ref{fig: Experiment_2}).

\subsection{Propagation of Joint-Level Realizability to Whole-Body Behaviors}
This experiment validates the proposed method on real hardware and examines how joint-level execution differences propagate to whole-body behaviors. VRA operates purely at the single-joint level, while the high-level whole-body RL controller is entirely unaware of joint position and velocity limits. All experiments are conducted on the same quadruped robot under identical motor and control settings, with joint boundaries enforced transparently by either VRA or VBAC.

\subsubsection{Two-leg Jumping Motion}
Two-leg jumping induces frequent joint boundary interactions through two distinct mechanisms: impact-driven excitation on stance legs at touchdown and command-driven excitation on swing legs during flight. Only joint position limits are imposed on selected stance- and swing-leg joints to intentionally activate boundary interactions. Both VRA and VBAC are evaluated under two time scales, $\{0.01\,\mathrm{s}, 0.05\,\mathrm{s}\}$.

As shown in Fig.~\ref{fig:Experiment_3}(a), at $0.01\,\mathrm{s}$ both methods execute the motion but exhibit boundary-induced oscillations. When the time scale increases to $0.05\,\mathrm{s}$, VBAC becomes unstable and fails, whereas VRA completes the motion with full constraint satisfaction and without noticeable oscillations, validating its robustness to reasoning–execution time-scale mismatch.

\subsubsection{Bipedal Standing with Upper-Limb End-Effector Motion}
Bipedal standing places stringent demands on joint velocities at the standing phase. In this experiment, both $0.01\,\mathrm{s}$ and $0.05\,\mathrm{s}$ settings are first evaluated, where VBAC fails at $0.05\,\mathrm{s}$~(Fig.~\ref{fig:Experiment_3}(b) and Fig.~\ref{fig: robot_result}(b)). Based on these results, each method is then operated under its best-performing time scale (VBAC: $0.01\,\mathrm{s}$, VRA: $0.05\,\mathrm{s}$) for bipedal standing with simultaneous upper-limb end-effector motion, which activates coupled position and velocity constraints across the whole body (Fig.~\ref{fig: robot_result}(c)).

Under these best-case configurations, VBAC induces strong oscillations in both lower-limb and upper-limb joints, leading to constraint-induced pattern changes and task failure. In contrast, VRA maintains smooth upper-limb motion and stable bipedal standing, completing the end-effector task. Taken together, these experiments validate that VRA serves as a missing acceleration interface that restores closure between kinematic acceleration reasoning and voltage-limited actuation.

%% file: content/Discussion.tex
\textbf{Implications for acceleration-based control.}
The results suggest that joint acceleration bounds should not be treated as purely kinematic quantities when robots operate close to actuator voltage limits. Many acceleration-based controllers use these bounds as an interface between high-level reasoning and low-level execution, implicitly assuming that admissible accelerations can be physically realized. Our experiments show that this assumption can fail near joint boundaries and high-speed regimes, where voltage constraints distort the acceleration realizability. VRA addresses this gap by exposing voltage-realizable acceleration bounds to upstream controllers, making actuator realizability part of the acceleration interface rather than a hidden low-level effect.

\textbf{Relation to whole-body and safety-critical controllers.}
VRA is complementary to whole-body, viability-based, and safety-critical control methods. It does not provide task-level feasibility or recursive feasibility guarantees by itself. Instead, it targets a lower-level assumption shared by these methods: that the selected joint acceleration can be executed by the voltage-constrained actuator. From this perspective, VRA can serve as an execution-aware layer beneath CBFs, hierarchical QPs, MPC, and viability-based filters, so that their reasoning is performed over acceleration bounds that better reflect the robot's physical capabilities.

\textbf{Limitations.}
(i) The current formulation is joint-level and interval-based, and therefore does not capture coupled multi-joint feasibility or whole-body contact constraints. (ii) It focuses on voltage-constrained actuation under the considered dq-axis motor model. The effects such as field weakening, thermal limits, and strong external loads are outside the current scope. (iii) Although residual-voltage compensation and experiments across motors and temperatures indicate robustness to moderate model mismatch, VRA still depends on calibrated actuator and joint parameters.

%% file: content/Conclusion.tex
This paper introduced VRA, a joint-level acceleration interface that connects discrete-time kinematic reasoning with voltage-limited actuation by grounding acceleration bounds in motor voltage constraints. Experiments on voltage-limited motors and a real wheel-legged quadruped robot demonstrate that VRA eliminates kinematically admissible but voltage-unrealizable acceleration commands, improving hardware execution near kinematic limits with reduced boundary-induced oscillations.

Future work will explore extending voltage realizability beyond interval-based bounds to richer feasible regions~\cite{Efficient_Paradigm,feasible_region} that capture multi-constraint interactions, and integrating voltage-realizable acceleration reasoning into safety-critical system-level control frameworks~\cite{Robust_Walking, CBF_Robust_WALK} for whole-body robots.

%% file: content/acknowledgement.tex
Lingwei Zhang would like to thank Kecheng Qin and Guangli Sun for their invaluable assistance in this work. This study was supported by the InnoHK initiative of the Innovation and Technology Commission of the Hong Kong Special Administrative Region Government via the Hong Kong Centre for Logistics Robotics.